%% file: main.tex
\documentclass[11pt]{article}

\usepackage[final]{acl}

\usepackage{times}
\usepackage{latexsym}

\usepackage[T1]{fontenc}
\usepackage[utf8]{inputenc}

\usepackage{microtype}

\usepackage{inconsolata}

\usepackage{graphicx}

\usepackage{lipsum}
\usepackage{tikz}
\usepackage{fancyvrb}
\usepackage{listings}
\usepackage{enumitem}
\usepackage{tcolorbox}
\usepackage[svgnames]{xcolor}
\usepackage{multicol}
\usepackage{siunitx}
\usepackage{subcaption}
\usepackage{subcaption}
\usepackage{colortbl}
\usepackage{CJKutf8}
\usepackage{booktabs}
\usepackage{float}
\usepackage[normalem]{ulem}
\usepackage{todonotes}
\usepackage{adjustbox}
\usepackage{array}
\usepackage{booktabs}

\usepackage[table]{xcolor}
\definecolor{tantable}{HTML}{DDD9C4}
\definecolor{bluetable}{HTML}{92CDDC}

\definecolor{pos1}{HTML}{D0E0F7} % Light Blue
\definecolor{pos2}{HTML}{9EC2EF} % Medium Blue
\definecolor{pos3}{HTML}{4A90E2} % Darker Blue
\definecolor{neg2}{HTML}{E68A00} % Dark Orange
\definecolor{neg2}{RGB}{235, 156, 92}  % below 45
\definecolor{secbg}{HTML}{D9D9D9} % Section Header Grey
\definecolor{green1}{HTML}{C6E0B4} % Light Green
\definecolor{green2}{HTML}{A9D08E} % Medium Green
\definecolor{green3}{HTML}{E2EFDA} % Very Light Green
\definecolor{lightgrey}{HTML}{E0E0E0} % Light Grey
\definecolor{darkgrey}{HTML}{A6A6A6}  % Dark Grey
\definecolor{secbg}{HTML}{D9D9D9}    % Section Header Grey

\definecolor{scorehigh}{HTML}{9EC2EF}
\definecolor{scoremid}{RGB}{222, 244, 214}  % 45--59.99
\definecolor{scorelow}{RGB}{235, 156, 92}  % below 45

\usepackage{booktabs}
\usepackage{longtable}
\usepackage{array}
\usepackage{makecell}

\newcommand{\ie}{{\textit{i.e.,} }}
\newcommand{\eg}{{\textit{e.g.,} }}

\title{Communicating with the Entire Population: Machine Translation of Emergency Alerts and Warnings}

\author{
    Jen Wilson\thanks{Corresponding author: \href{mailto:jenwils@uw.edu}{jenwils@uw.edu}}, {Vipasha Bansal}, {William D. Lewis} \\
    Computational Linguistics Program \\ University of Washington \\
    }

\begin{document}
\maketitle
\begin{abstract}
It is essential that we communicate with the entirety of an affected population during a crisis, including those who do not speak a region's dominant language(s). Commercial Translation tools offer a tempting solution to broaden access and can provide a lifeline when no other resources are available. However, Emergency response managers often operate in the dark regarding the quality and adequacy of these automated outputs. In this paper, we conduct a nationwide survey of emergency managers to understand their current workflows, perspectives, and operational challenges. We then evaluate commercial translation tools against these needs, shedding light on their utility and offering practical guidance on how and when these tools can be safely used in emergency response. 
\end{abstract}

\section{Introduction}
\label{sec:intro}
\input{Sections/Introduction}

\section{Wireless Emergency Alerts}
\label{sec:WEA}
\input{Sections/WEAs}

\section{Establishing the Need: A Survey of Emergency Managers}
\label{sec:need}
\input{Sections/Need}

\section{Differences between MT and LLM use by Emergency Managers}
\label{sec:difference}
\input{Sections/Differences}

\section{Quality of Commercial MT Tools in Crisis Response}
\label{sec:quality}
\input{Sections/Quality}

\section{What to Do and What Not to Do with Commercial MT Tools}
\label{sec:workflow}
\input{Sections/Workflow}

\section{Conclusion}
\label{sec:conclusion}
\input{Sections/Conclusion}

\section*{Limitations}
\label{sec:limitations}
\input{Sections/Limitations}

\section*{Ethical Considerations}
\label{sec:ethics}
\input{Sections/Ethics}

\section*{Acknowledgments}

This work was partially supported by the National Science Foundation under the award CIRC 2346335. We would like to thank the two LT4CPR workshop reviewers for their very helpful comments and suggestions, all of which helped improve this paper. 
We also extend our warmest THANK YOU to the members of LAWWW, whose contributions to the survey, including their suggestions, questions, and support, made this paper what it is today. Without LAWWW, this paper could not have been written.

\bibliography{bib/crisis.bib, bib/aacl}

% \clearpage
\section*{APPENDIX}
\label{sec:appendix}
\input{Sections/Appendices}

\end{document}

%% file: Sections/Introduction.tex
Communication during emergencies is a core part of public safety. Access to timely information during a crisis is as essential as access to "food, water, shelter, and medication” \citep{Meier2015}. Emergency alerts and warnings systems, such as Wireless Emergency Alerts (WEA)\footnote{\label{fn:WEAs}https://www.fema.gov/emergency-managers/practitioners/integrated-public-alert-warning-system/public/wireless-emergency-alerts}—must reach the \textit{entirety} of an affected population, rather than an English-fluent subset. Failing to achieve equitable coverage puts lives directly at risk.

This outreach is difficult in practice. Local emergency managers, public information managers, and communication staff often serve multilingual communities while working with limited time, staffing and budgets, sometimes across dozens or even hundreds of languages \cite{Ciribuco-et-al-multilingual-crises-2024}. In regions where alerts are distributed only in the dominant language(s), non-fluent residents are inevitably excluded, which can have devastating consequences in time-sensitive crises \citep{Villarreal-et-al-2025-barriers, LinguisticChallengesinSevereWeatherCommunicationComparingEnglishandSpanishSpeakingPopulations, obrien-federici-2019-crisis-translation-needs}.

Because human translation is often cost-prohibitive or too slow for local jurisdictions, practitioners increasingly turn to automated solutions. Commercial Machine Translation (MT) systems, such as Google Translate\footnote{\label{fn:google}https://translate.google.com/} and Microsoft Translator\footnote{\label{fn:microsoft}https://www.microsoft.com/en-us/translator/}, are widely available and support a large number of languages. Large Language Models (LLMs) such as ChatGPT\footnote{\label{fn:chatgpt}https://chatgpt.com/}, Claude\footnote{https://claude.com/}, and Google Gemini\footnote{\label{fn:gemini}https://gemini.google.com/app} are also increasingly used for translation.

These tools may offer speed and convenience, but emergency messaging demands accuracy and usability. Inaccurate translations risk delivering alerts that are uninterpretable or dangerously misleading. Crucially, strong MT performance in one domain or language pair does not guarantee cross-lingual or cross-domain reliability \citep{bogoychev-chen-2021-highs, zouhar-etal-2024-fine}, requiring comprehensive multi-target evaluation to ensure safety.

Although this paper focuses on emergency information translation in the United States, the challenges we describe are not unique to any one country and are likely familiar to officials in many multilingual settings. As claims regarding Artificial Intelligence (AI)\footnote{Machine Translation being one of many forms of AI.} expand across crisis management, practitioners face a rapidly evolving landscape of software tools and vendor claims \citep{roleofAIinEM, centaurs}. In this paper, we seek to shed light on automated translation in emergency response from two angles.\footnote{In this work, we chose to focus on the quality and utility of commercially available tools that do translation. We leave the analysis of MT workflows, logistics of their implementation and use in the field, etc., to others.} First, we survey emergency managers across the United States to understand current translation practices, needs, and challenges. 

Second, we examine the translation tools available to practitioners and review existing research on their evaluation and use in crisis settings. By bringing together the perspectives of emergency management practitioners and research on translation technologies, we hope to help bridge the gap between the development and evaluation of these tools and the contexts in which they are used.

%% file: Sections/WEAs.tex
In the United States, the Wireless Emergency Alerts (WEA)\footref{fn:WEAs} system broadcasts alerts to mobile and other WEA-enabled devices, providing actionable information in times of crisis. However, native support is restricted to English, leaving non-English speakers at a disadvantage unless they proactively sign up for translations.

In 2023, the US Federal Communications Commission \cite{wea_2023} recognized that ``it is essential that the public be able to receive in accessible language and format WEA Messages that are intended for them...to help the millions of people with access and functional needs, including people who primarily speak a language other than English or Spanish...better understand and take protective actions.'' Despite this, the ability of agencies to meet multilingual and accessible alerting needs varies widely depending on local demographics, staffing, funding, and available translation infrastructure.

%% file: Sections/Need.tex
The role of language technology in emergency management is underexplored and deserves attention. \citet{roleofAIinEM} surveyed emergency managers to identify general perceptions and potential use cases for emerging AI. The specific adoption and efficacy of automated translation tools in crisis contexts require dedicated study.

To examine current translation practices, needs, and operational hurdles, we surveyed members of the Language \& Accessibility for Alert \& Warning Workgroup (LAAWW)\footnote{\url{https://sites.google.com/view/laaww}} in December 2025. LAAWW members represent 42 U.S. states, territories, and Tribal Nations who are "working toward greater language and disability access in emergency alert and warning systems."

We used purposive sampling strategy to reach practitioners actively involved in language access.\footnote{This approach surfaces qualitative operational insights; future work with randomized sampling would be required for broader statistical generalization.} Therefore, this survey is not intended to be statistically representative of all U.S. emergency management agencies; rather, it provides insight into the tools, decision-making processes, and challenges involved in producing multilingual alerts. Survey details are in Appendix \ref{app:survey_questions}.

The survey asked about respondents' roles and responsibilities, languages spoken in their communities, translation tools and practices, and challenges encountered in multilingual communication. 
We received 25 responses, approximately 25\% of meeting attendees.\footnote{Attendance at LAAWW meetings is approximately 100 people per month, representing approximately 80 unique entities (agencies, vendors, cities, counties, etc).
%, though specific attendance for this meeting is unavailable.
}

More than half of respondents identified themselves as a \textit{Local Emergency Manager} or \textit{Public Information Officer}, while 36\% identified as an \textit{Operations, Logistics, or Communications Manager} or \textit{Language Coordinator or Interpreter}.\footnote{The survey allowed respondents to select more than one role.}

Participants noted that while many jurisdictions have residents who speak a variety of languages, only a subset of these languages are consistently supported for time-critical alerts. Agencies must prioritize languages while balancing accuracy, usability, speed, and cost. Translations should be usable and accurate, while also affordably and quickly produced. 

Although participants prioritized quality translations, for those who shared their challenges, 100\% cited money, time, and/or limited resources.

\subsection{Translation Needs Across LAAWW Membership}

Translation needs vary substantially across jurisdictions represented in the survey. Respondents reported serving communities where between two and more than 200 languages are spoken. English was the primary language across all reported jurisdictions and Spanish was the most consistently supported additional language. However, the remaining language needs varied considerably by region. A partial list of languages that survey participants work with is available in Appendix \ref{tab:laaww_languages}.

The scale of these differences creates distinct operational requirements. Smaller jurisdictions may be able to reach most residents by translating alerts into only a few languages, while larger and more linguistically diverse jurisdictions may need to support many more. For example, one respondent from Maryland supports a community of approximately 5,500 people and three commonly spoken languages. Another works in New York City, where more than 200 languages are spoken across a population of more than 8.5 million. While we can look at average language populations, nearly every area has populations that deviate from that national average. These jurisdictions face fundamentally different translation-planning challenges.\footnote{Appendix \ref{app:languages} displays this data as percentage off expected.} Survey respondents support over 36 million people across the United States\footnote{The population information reported here reflects the total population of respondents’ jurisdictions and do not imply that all residents require translated alerts or that language services are uniformly provided across all languages in each jurisdiction. Rather, they illustrate the scale and diversity of contexts in which respondents operate.}. 

Agencies must identify the languages most important for their communities and develop processes for producing translations that are accurate, usable, affordable, and fast enough for time-critical alerts. The central need is not simply translation into more languages, but scalable and locally informed translation capacity. For more details on translation needs and speaker populations, as reported by the LAAWW membership, please see Appendix \ref{app:more_on_translation_needs}. 

\subsection{Translation Tools and Practices}

\begin{figure}
    \centering
    \includegraphics[width=1\linewidth]{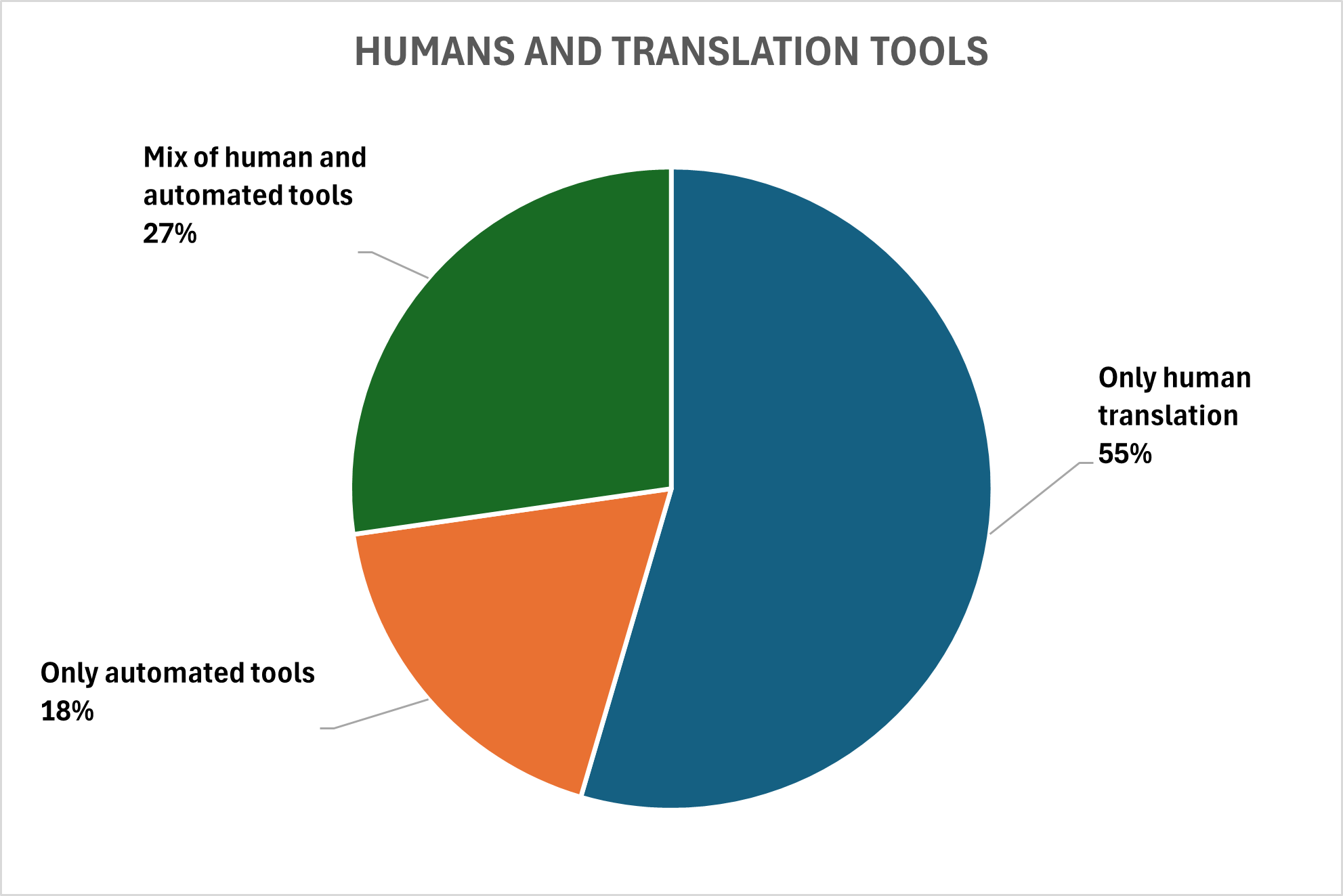}
    \caption{Human Translation and Translation Tools}
    \label{fig:humans_and_tools}
\end{figure}

The majority of respondents  (55\%) relied exclusively on human translation and did not use automated tools. Another 27\% used a mix of human translators and automated tools, while 18\% relied exclusively on automated tools (Figure \ref{fig:humans_and_tools}). For those using human translation, work was performed by organization staff (45\%), community members or volunteers (26\%), and contractors or commercial vendors (29\%). These results show that jurisdictions use a range of staffing models to produce multilingual alerts. 

Among respondents using translation-specific commercial tools, Google Translate\footref{fn:google} was the choice in 90\% of cases, followed by DeepL Translator\footnote{\url{https://www.deepl.com/en/translator}} at 20\%.\footnote{\label{fn:multiple_responses}
Respondents could specify more than one tool and the percentages do not necessarily add to 100\%.} A minority (and we suspect growing) number of respondents use LLM–based tools, with ChatGPT\footref{fn:chatgpt} used in 30\%
of cases and Google Gemini\footref{fn:gemini} used in 10\% of cases.\footref{fn:multiple_responses}
Given the small sample size, reported use of LLM-based tools likely reflects exploratory or supplemental use rather than standardized, organization-wide deployment.
% \todo{check this quote}
These results are consistent with broader interest in AI among emergency managers. \citet{roleofAIinEM} found that 75\% of their respondents were interested in using AI for public and stakeholder messaging, including multilingual translations. Together, these findings suggest that automated tools are already part of some emergency communication workflows, while their organizational adoption remains uneven.

\subsection{Factors Influencing Tool Selection}
\label{sec:choice}

Respondents selected translation tools primarily on the basis of practical considerations rather than formal requirements. Availability or ease of access was a criteria for 56\% of respondents, and perceived accuracy or quality was important for 52\%. Cost or budget constraints influenced 40\% of respondents' choices, while ease of use or staff comfort influenced 32\%. These results suggest that agencies favor tools that are accessible, affordable, and sufficiently reliable for everyday workflows.

Compliance requirements (16\%) and "other" factors (12\%) were cited less often, indicating that regulatory or external mandates play a smaller role in shaping tool choices. 

\begin{figure}
    \centering
    \includegraphics[width=1\linewidth]{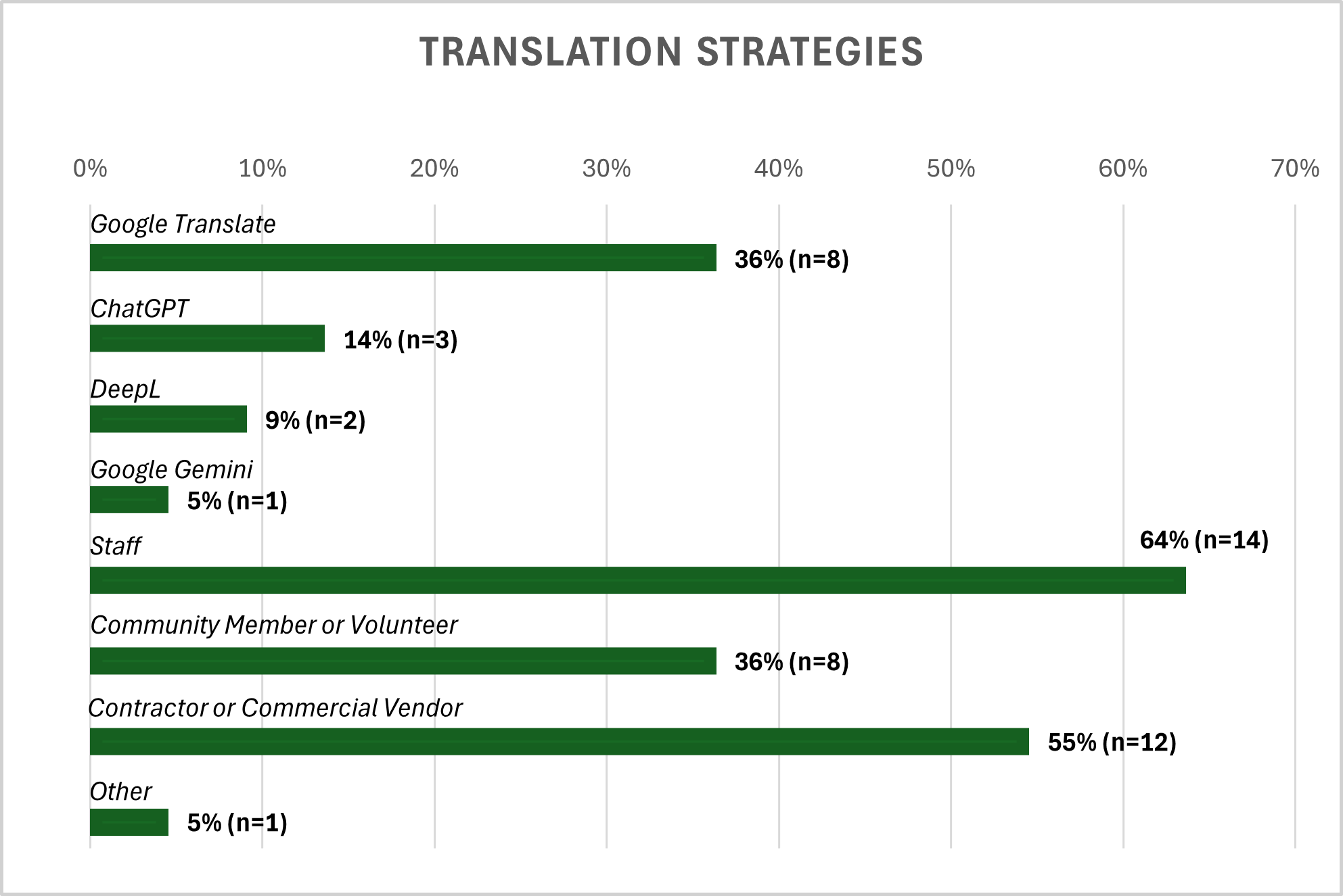}
    \caption{Strategies of respondents to obtain translations.}
    \label{fig:strategies}
\end{figure}

Tool adoption therefore appears to be driven more by immediacy and feasibility than by formal risk assessment or regulatory guidance (Figure \ref{fig:strategies}). This pattern is consequential because a tool that is convenient to access may still require systematic evaluation before it is used for high-stakes messages

\subsection{Evaluation and Review Practices}

Formal evaluation of translation tool output was uncommon. Only 10\% of respondents reported having a formal process for assessing accuracy, usability, or effectiveness. Sixty percent used an informal evaluation process, while 30\% reported having no evaluation process.

Processes for reviewing or updating tools were similarly inconsistent. Forty percent of respondents reported reviewing or updating tools as needed, 25\% reported having no formal review process and another 25\% were unsure whether such a process existed. The remaining 10\% reported that their tools were not reviewed at all. Thus, many jurisdictions may be using translation tools in high-stakes contexts without systematic procedures for evaluating or monitoring their performance, creating a risk of uneven message quality across languages.

\subsection{Operational Challenges}
Respondents described multiple barriers to producing multilingual alerts, including limited time, funding, staff, and management support.

Translator time and availability were also recurring concerns. Participant \colorbox{LightGreen}{L1} emphasized the need to ``minimiz[e] time delays in alerts and warnings...especially with translations into various languages.'' Participant \colorbox{LightCyan}{P1} similarly described the ``challenge [with] human translation is always time and availability'' of translators. 

Participant \colorbox{LightPink}{H1} reported that their state maintained a registry of approved interpreters, but very few were available in their rural county. As linguistic diversity increases, agencies may need to support additional languages even when human translators and external vendors are slow or unavailable.

Respondents also reported that the usefulness of translation tools varied by language. Participant \colorbox{MediumTurquoise}{R1} explained ``multiple languages in the state are very unique thus singular translation services for only a few languages [have] proven ineffective in [the] past.'' This aligns with our CHRF++ findings (Section \ref{sec:quality}), which showed significant variation across languages and systems

Beyond operational barriers, participants also noted that language access does not guarantee message comprehension, particularly for populations with limited literacy or differing communication preferences. These constraints show a deeper challenge: language access alone is insufficient. The findings point to the need for complementary accessibility strategies, such as plain-language messaging, audio or pictorial alerts, and community-based dissemination, especially in high-risk populations \cite{Villarreal-et-al-2025-barriers}.

\begin{figure}
    \centering
    \includegraphics[width=1\linewidth]{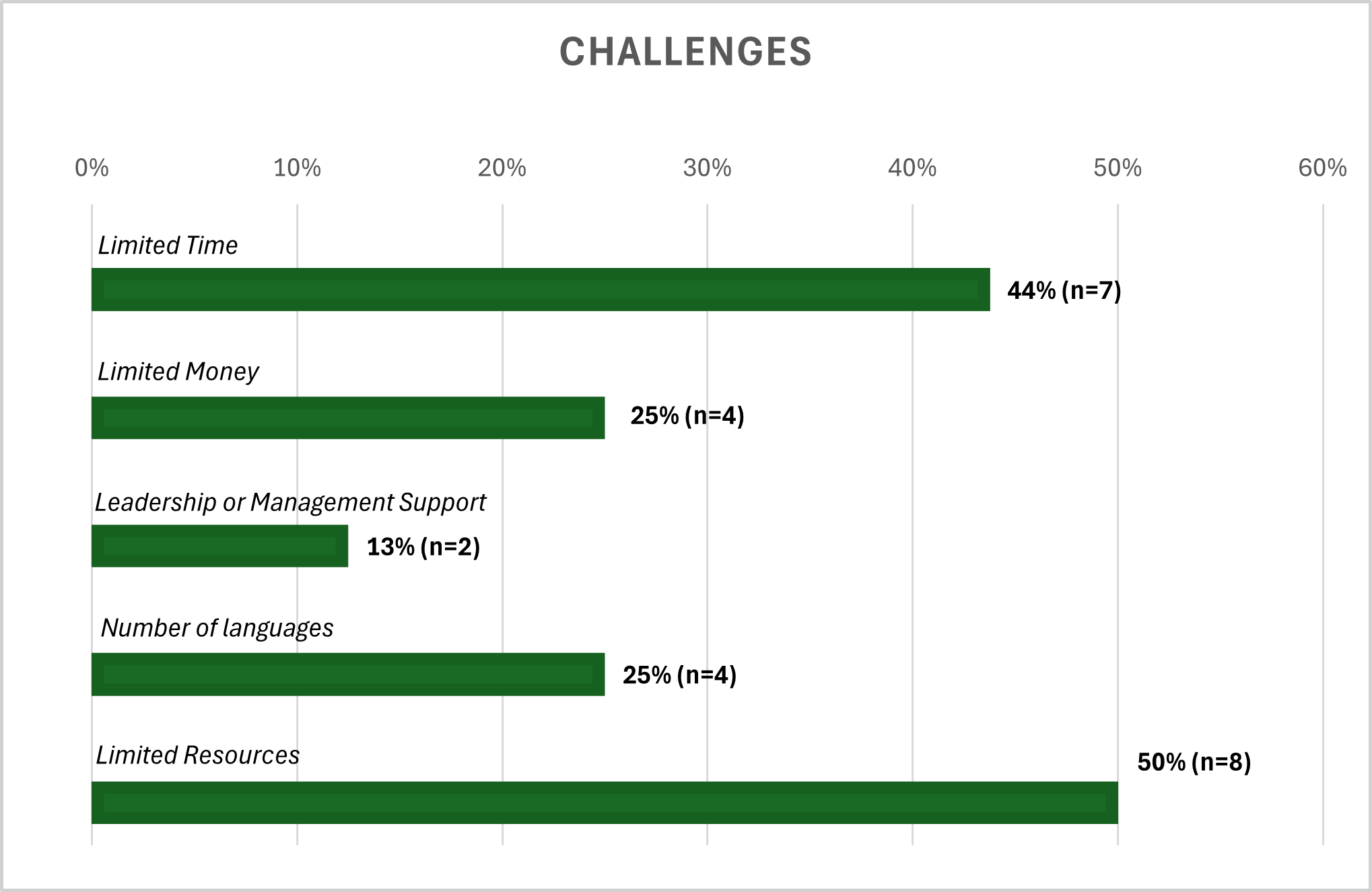}
    \caption{Operational challenges faced by survey respondents.}
    \label{fig:challenges}
\end{figure}

%% file: Sections/Differences.tex
Our survey demonstrated that respondents in emergency management and communications use both MT and LLM-based tools. There are many differences between the two technologies, but for the purposes of this paper, we will discuss the expected outputs and the types of data used in the training of Google Translate and OpenAI's ChatGPT.\footnote{Discussion of other similarities and differences, such as model architecture, training objectives, training process, inference mechanisms, etc. is beyond the scope of this paper.}

MT is traditionally trained on a corpus of aligned sentence pairs \citep{buildingmachinetranslationsystems}, where the sentences corresponding to one another in each language have the same meaning.
%with both sentences having the same meaning but in two different languages. 
The MT models learn how text in one language corresponds to equivalent text in another language. Commercial tools, such as Google Translate and Microsoft Translator, are trained in this manner and publish lists of languages that are supported by their services.\footnote{For a list of languages supported by Google Translate, see: https://docs.cloud.google.com/translate/docs/languages. For Microsoft Translator, see: https://www.microsoft.com/en-us/translator/languages/.}

While LLMs such as ChatGPT are pre-trained on publicly available data (websites, news, etc.), multimodal data, non-English data, and data licensed for training \citep{openai2024gpt4}, the specific data sources are rarely disclosed. The data may also include aligned sentence pairs and text in multiple languages \citep{chua2025crosslingualcapabilitiesknowledgebarriers}. However, unlike the MT services discussed in the previous paragraph, LLMs generally do not list what languages they support for MT. In fact, MT may arise as an emergent property rather than as a training goal \cite{wei2022emergentabilitieslargelanguage}. LLMs are trained to generate plausible output, and will provide output even when lacking sufficient data to provide one that is correct. This can lead ``models to convert uncertainty into content that is often plausible but wrong'' \citep{kalai2026evaluating}.

%% file: Sections/Quality.tex
The survey of LAAWW membership shows a clear interest in automated tools for translation. Several studies have evaluated the use and performance of MT in crisis contexts, such as for emergencies~\citep{mtinemergencies}, multilingual crisis-related communication~\citep{mtincrisis}, or pandemic-related  communication across multiple languages~\citep{anastasopoulos-etal-2020-tico}. More recent work has evaluated both MT systems and LLMs on emergency-response texts and multilingual crisis data, finding limitations in areas including domain-specific terminology, contextual ambiguity, register, and the preservation of urgency \citep{staiano-etal-2025-italert, Ticona_Anastasopoulos_2026, court-etal-2026-llms}.  Less attention has been given to systematically comparing currently available commercial translation tools on emergency alert content across multiple languages. This distinction is important for emergency managers, who are more likely to encounter translation tools as commercially available services than as models requiring specialized technical expertise to use or adapt.

\subsection{Quantitative Evaluation}
\citet{bansal-etal-2026} evaluated seven commercially available MT systems on the TICO-19 pandemic-related benchmark dataset \citep{anastasopoulos-etal-2020-tico}, which contains English-language COVID-19 material translated into 36 languages, and assessed translation usability based on whether the output was sufficiently understandable and adequately conveyed the relevant information. Their results demonstrate substantial variation in translation performance across languages and systems, particularly for lower-resource languages.

\citeauthor{pang2025} (\citeyear{pang2025}) find that LLMs trained across multiple domains perform consistently across a range of tasks but can fall short of domain-specific models. To investigate whether this pattern extends to crisis communication, we compare scores on the full TICO-19 dataset with scores on its News and Announcements (N+A) subset, whose short, information-rich texts more closely resemble Wireless Emergency Alerts (WEAs). Because WEAs are issued in English, we focus on English-to-target-language (EX) translation scores.

LAAWW members identified 24 supported languages that overlap with the languages represented in TICO-19. This overlap enables a unique domain-specific evaluation of emergency-relevant text for those languages, allowing a comparison with general-purpose content.  Following \citeauthor{anastasopoulos-etal-2020-tico}'s (\citeyear{anastasopoulos-etal-2020-tico}) categorization, we group the languages into three tiers: \textit{Pivot} (high-resource), \textit{Important} (mid-resource), and \textit{Priority} (low-resource).

We examine CHRF++ scores for both the overall and N+A subset of TICO-19. CHRF++ is a quality metric widely used in the MT field for measuring quality~\cite{Popvic-CHRFPP-2017}. The differences between these scores for some language-system combinations, suggests that performance on a broader evaluation set does not necessarily predict performance on text with characteristics more similar to emergency alerts. 

Pivot languages generally achieved the strongest Overall scores, while Important and Priority languages showed lower and more variable performance. No system was consistently strongest across languages, and the best-performing system sometimes changed between the Overall and N+A evaluations. These scores suggest that overall benchmark scores are insufficient for predicting performance on emergency-relevant text and that evaluation should account for both language and domain.\footnote{See Appendix \ref{app:evaluation_results}, Table~\ref{tab:chrf_scores_spaced} for full scores.}

For example, Microsoft's Hindi N+A CHRF++ score was 13.16 points \textit{lower} than its Overall score, Claude was 12.24 points lower, and GPT-5.4-mini was 10.33 points lower. In fact, all of the Pivot languages showed lower scores for the N+A subset, except for Swahili, for which all services were between 1.54 and 3.13 \textit{higher} (Appendix \ref{app:evaluation_results}, Table \ref{tab:chrf_diff_scores_colored}). For most language--system pairs, CHRF++ scores were higher on Overall than on the N+A subset, with particularly large differences for Hindi and Arabic. However, this pattern was not universal: Bengali, Khmer, Kinyarwanda, Swahili, Somali, Tamil, and Zulu had lower Overall than N+A scores for at least one system.

\subsection{Qualitative Evaluation}
\label{sec:QualitativeEval}

Qualitative analyses by \citeauthor{hajek-et-al-2024} (\citeyear{hajek-et-al-2024}) identify errors that may sound fluent, but substantially alter the intended meaning of emergency instructions. For example, the English instruction “Try to position the car \textit{towards} the approaching fire” was translated to Chinese and then back into English as “Try to park your car \textit{near} the fire,” reversing the intended instruction and potentially creating a very serious safety risk.\footnote{They did not note which system produced this translation.
\citet{hajek-et-al-2024} gathered their translations in 2023. Upon subsequently examining Google Translate (captured on 14 Jan, 2026), the Chinese translation is \begin{CJK}{UTF8}{min}
试着将汽车开向逼近的火焰。 
\end{CJK}}

\citeauthor{hajek-et-al-2024} (\citeyear{hajek-et-al-2024})  also describe problems with lexical ambiguity. The instruction ``shelter in a room'' was translated into Spanish as ``refugio en una habitaci\'on,'' which when backtranslated, means ``a refuge in a room.'' The word \textit{shelter} in this case was translated as a noun (\textit{refugio}), when actually it was a verb (a command to take shelter). It may be tempting to assume that \textit{something} is better than \textit{nothing} when it comes to emergency messaging. However, these examples demonstrate that that is not always the case. 

These examples illustrate a type of MT failure that is dangerous in emergency contexts: mistranslations that are fluent, possibly plausible-sounding are therefore unlikely to be flagged by non-expert reviewers, but convey meaningfully different instructions \citep{sorak_2024}.

Additionally, \citet{Ticona_Anastasopoulos_2026} demonstrate that LLMs inconsistently translate urgency, such as ``a gas leak scenario classified as Critical in English is downgraded to Low in Marathi and Not Urgent in Amharic.'' 

Qualitative evaluations provide examples that show where MT has gone wrong and can be useful in providing actionable content that can help guide how translations are done, used, or evaluated.
However, they do not give an overall picture of MT output but only specific instances identified by human translators and may be, or may not be, representative of broader issues. This limitation underscores the need for domain-specific and systematic evaluation frameworks for MT in emergency settings.

%% file: Sections/Workflow.tex
The gold standard for translations of emergency alerts and messaging is to translate all anticipated messages by a human before they are needed \citep{Federici02102022,Villarreal-et-al-2025-barriers}. 
When an organization has limited resources, or an unanticipated crisis occurs, or where professional translators are not available, this process may not be possible. Commercial machine translation can bridge the gap and get information out to people quickly, while buying time to have a more thorough translation prepared.

Based on work by
 \citet{Ticona_Anastasopoulos_2026} and \citet{bansal-etal-2026}, we can see that there are gaps in the quality of commercial and open-source MT tools in the crisis domain cross-lingually, and potentially undesirable errors in output, such as those noted by \citet{hajek-et-al-2024}. Knowing the quality gaps can help emergency managers integrate translation tools into their workflows appropriately. It is important to know how one can improve the quality of MT and how much a human should be in the loop. In this section, we will look at ideas and rules of thumb that emergency managers can adopt for using such tools.

\subsection{The Epic Highs and Lows of Raw Machine Translation}

As MT quality has improved, especially for high resource languages, the use of raw MT has become more prevalent, especially in contexts where human translators are not available \citep{Nurminen-2021-phd,nurminen-papula-2018-gist,kaspere-etal-2021-sustainable,vieira-etal-2022-machine,kenny-etal-2022-machine}—a constraint reflecting the standard operating environment of many LAAWW practitioners.

However, there is a significant literature on the use of MT in crises, with the general recommendation is that humans should be in the loop to prevent severe errors with potentially catastrophic consequences\footnote{We discuss notable examples in Section \ref{sec:QualitativeEval}.} \citep{cadwell-etal-2019-more-than-tweets,obrien-federici-2019-crisis-translation-needs,federici-cadwell-training-citizen-translators,Federici02102022}. Reflecting these risks, many Language Access Policies discourage the use of raw MT, \ie using the direct output of MT engines without any post-editing or human-in-the-loop \citep{la_language_policy, digitalIntroductionTranslation, title9workinggroup}.  

Strictly avoiding raw MT may not be feasible when contending with budget limits, as highlighted by our survey respondents. But there are things that can be done to help mitigate some of these dangers, either through simplifying text, preparing messages in advance, use of templates, or doing as much as possible in advance of crises.

\subsection{Translated Templates}
\label{sec:templates}

The Federal Communication Commission (FCC) provides templates for WEAs, translated into the 13 most commonly spoken languages in the United States.\footnote{
https://www.fcc.gov/multilingual-wireless-emergency-alerts hosts example templates.} 
By using templates like these, one can sidestep the use of MT altogether in certain crisis scenarios.

We recommend creating or repuposing templates in advance of crises that may occur regularly. Use the following steps to create templates that can be repurposed:
\begin{enumerate}
\item
%\textbf{(1)}
\textbf{Pre-approval:} Draft templates, using short, plain-language messages for hazards or actions. 
\item 
%\textbf{(2)}
\textbf{Placeholder management:} Use clearly labeled and unambiguous placeholders in the templates for event specific details (\eg location and place names, dates, times, etc.). Ensure the placeholders are untranslated, and have the correct placement for the target languages. 
\item 
%\textbf{(3)} 
\textbf{Rapid deployment:} When an incident occurs and a message template has already been prepared, it can be filled in and translated to additional languages through automated tools.
\end{enumerate}

\subsection{Text Simplification}
\label{sec:simplification}

Text simplification has long been used to improve MT output, and is discussed in the literature under various guises: text simplification \citep{stajner-popovic-2016-text}, controlled language \citep{aikawa-etal-2007-controlled-language}, and pre-editing \citep{sanchez-gijon-kenny-2022-pre-editing}, among others. In the distributed messages setting, such as with WEAs, all messages should be as clear and direct as possible to increase the chance of information being translated accurately.

\citeauthor{stajner-popovic-2016-text} (\citeyear{stajner-popovic-2016-text}) find that text simplification can improve the fluency of the MT output. While their work focuses on automated simplification, the same principle can be extended to include manual simplification techniques, \ie where the authors of the messages simplify the text themselves. \citeauthor{hajek-et-al-2024} (\citeyear{hajek-et-al-2024}), for example, recommend avoiding culture-specific terms, avoiding two-part verbs (\eg look up, get along with, etc.), and avoiding elements with multiple grammatical functions. Recommendations to simplify crisis communications before translation have also been made by local governments \citep{kingcounty}, policy reports \citep{fordham}, and government reports \citep{hajek-et-al-2024}.

LLMs can also be used in text simplification \citep{guidroz-etal-2025-LLM-text-simplification}. For example, the following text could be used as a prompt to an LLM to simplify text supplied by the user:

\textit{Rewrite the following text in plain language at approximately a 6th-grade reading level. Use short sentences and everyday vocabulary. Keep all safety instructions, locations, and times accurate}.

The text simplification process should be treated as a separate step from translation. Evaluating the simplified text before translation can help confirm that the source message remains accurate, retains all essential information, and is adequately simplified and ready for the next stage. Separating these steps makes it possible to identify and correct problems with the source message before it is disseminated to the public.

\subsection{Using LLMs for Translations}

 The majority of LAAWW members use human translation (Figure \ref{fig:humans_and_tools}), but respondents also indicated interest in translating with LLMs. The use of general-purpose LLMs, such as GPT and Gemini, in high-stakes communication contexts introduces risks that go beyond translation quality as measured by evaluation metrics. LLMs are known to produce fluent but semantically incorrect outputs — a phenomenon called ``hallucination'' — at rates that are difficult to predict and vary by language, domain, and prompt formulation \cite{ji-surveyofNLhallucination}. In crisis communication, this unpredictability is a serious concern.

To reduce the risk of hallucinations, we recommend using a prompt such as the one below, which instructs the LLM to leave text untranslated when it is uncertain. Although this prompt may lower the likelihood of hallucinations, it is not guaranteed. For languages not well supported by the model, LLM-based translation should be avoided. See Appendix \ref{app:evaluation_results} for additional guidance.

\textit{Translate this message into [TARGET LANGUAGE]. Be literal. Do not add, omit or invent information. If a phrase is ambiguous, mark it as [UNCERTAIN] and leave it untranslated. Preserve names, dates, and numbers exactly. After the translation, provide a one-line confidence: High / Medium / Low and list any parts you were unsure about.}

The authors acknowledge the mixed messaging this section provides. While we suggest that LLMs not be used for translations, we also give guidance on how to use LLMs for translations. As researchers, we are hesitant to recommend the use of LLMs for the reasons stated above. As people who have some understanding of the community that uses these tools, we know that LLMs may be used for translations regardless of what an academic paper may recommend. To that end, we also recommend strategies to use LLMs as effectively as possible.

The models used in commercial LLMs can change with little notice to end users, making it difficult to track how system behavior evolves over time. Because these system updates may be undocumented or only partially disclosed, a prompt or output that is usable at one point in time may produce different results later, even when the input remains unchanged. For emergency management practitioners who use LLMs for translation, these changes can be especially consequential. Small shifts in wording, terminology, or tone may affect whether a message is interpreted accurately and in a way that supports timely response. This potential inconsistency in output should be considered if choosing to use an LLM instead of a commerial MT tool, such as Google Translate or DeepL.

\subsection{Post-Editing and Usage}

It is challenging for organizations to determine how to best allocate limited resources. One option for translation, beyond the methods discussed previously (\eg using templates, simplification, etc.) is to spend  resources on post-editing rather than on message creation and translation. In this scenario, human translators rely on MT to generate a translated output, and then subsequently correct any errors in the output.   This allows translators to spend their time reviewing and editing translated messages rather than translating messages from scratch. The use of MT in workflows has been shown to reduce costs.

To make messages broadly consumable, in addition to the steps outlined in Sections \ref{sec:templates} and \ref{sec:simplification}, we recommend that the messages are shared on available public and social platforms \citep{engage_and_mobilize}. Some platforms provide built-in translation features, allowing users to translate messages without copying the text into a separate program or application. However, this approach removes the ability to monitor quality of the translations and increases the risk that people will get misleading, incorrect, or harmful information. It will be up to the user to determine whether this will work for their situation.

MT engines can produce errors, some of which can have catastrophic consequences, such as examples from \citet{hajek-et-al-2024}. Although quality evaluation is certainly possible, it is useful to determine \textit{adequacy} for the specific need versus absolute \textit{quality}. In other words, a translation that has some limited errors or is ungrammatical may still be of sufficient quality and good enough to be used. For suggestions on how to determine the adequacy of translations, see Appendix \ref{appendix:more_on_evaluations}.

Ad hoc translations likely occur regularly during crisis situations, given the ubiquity of translation tools and people's familiarity with them. In urgent situations, messages are often issued first in the dominant language(s) and translated for dissemination in other languages later. For those who do not speak the dominant language,  urgency will drive them to quickly obtain their own translations, using some of the same commercial translation tools described in this paper. In some cases, a translated message may not be disseminated in a specific language at all, forcing recipients to translate the message themselves. Organizations should therefore consider not only the quality of translations \textit{they} produce, but also the translations that recipients are likely to obtain independently when official translations are unavailable.

%% file: Sections/Conclusion.tex
We recommend having humans in the loop in translating, editing, and evaluation whenever resources permit. In practice, however, jurisdictions that communicate in dozens of languages may not be able to review each message by hand. Automated workflows can expand translation capacity, but should be designed around the risks of the message, language, and tool rather than treated as a universal replacement for human review.

An important point to make is that MT engines will make errors, and not all errors are created equally. The types and severity of errors can vary greatly depending on the content, \ie the complexity of the original text or the amount of resources for the target language. As we have shown, MT tends to do better with higher-resource languages such as Indonesian or Spanish than for lower-resource languages such as Amharic or Pashto. Performance on the N+A subset did not always follow performance on the broader evaluation. These differences make language- and domain-specific evaluation essential.

In a world where communities are becoming ever more multilingual, translation will remain an important component of emergency and crisis communication. Fortunately, the quality and utility of MT is improving, but its use should be paired with realistic expectations, along with human oversight. In crisis settings, the goal is not simply to produce a translation, but rather to deliver information that affected communities can understand and act on safely.

%% file: Sections/Limitations.tex
The survey done for this paper captures the languages that US-based agencies \textit{report} that they need to support. It does not systematically compare those responses against actual distribution of non-English speakers in the respondents' jurisdictions as measured by US Census data or available American Community Survey data.

%% file: Sections/Ethics.tex
Considerable care was taken in designing and administering the survey to ensure that the questions were appropriate to the study objectives, respectful of participants, and attentive to the needs and perspectives of the communities being studied. The survey was developed after careful consideration of the roles and responsibilities of LAAWW members. The methodologies and goals of the study were presented during a LAAWW meeting and members were able to ask questions. This process helped ensure that the survey questions, framing, and proposed use of the data were clearly understood and acceptable to the participating community.

The study was also submitted for review through the University of Washington (Seattle) Human Subjects Division Institutional Review Board (IRB). The IRB determined that the proposed activity constituted human subjects research and qualified for exempt status. Data collection proceeded in accordance with the requirements and conditions of that determination.

Participation in the survey was designed to protect participants' privacy. Participants were able to respond anonymously and were not required to provide identifying information. Participants were given the option to voluntarily provide contact information and geographic location if they wished to do so. These fields were optional and were not required for participation. The study therefore collected only information that participants chose to provide, with identifying information separated from the anonymous survey responses as appropriate for the purposes of analysis and reporting.

Taken together, these steps were intended to ensure that the survey process was conducted thoughtfully and responsibly, from initial community consultation through IRB review, data collection, privacy protection, and subsequent analysis.

%% file: Sections/Appendices.tex
\setcounter{section}{0}
\renewcommand{\thesection}{\Alph{section}}

\section{Translation Dataset}
As part of our collaboration with LAAWW, we are compiling a collection of pre-translated emergency alerts from multiple agencies across the United States. The alerts span various emergency types, including (\eg weather, public health, hazmat, etc.). Each alert includes the original English message, professional human translations into dozens of languages, and metadata such as agency, date issued, language pair, alert type, etc.

We intend to make this dataset available to all those in the emergency management community who wish to use these pre-translated alerts. We also intend to make the dataset available to the research community to support future work on translation quality evaluation, multilingual emergency communication, etc. We are actively growing this data set. If you have data to share, please contact the authors.

\section{Participant Information}
Table \ref{tab:participant_information} is intended to give the reader further insight into participant comments. Only survey participants who are quoted in this paper are included in this table.

\input{tables/respondants}
\section{Translation Needs and Speaker Populations}
\label{app:more_on_translation_needs}

\begin{figure}[H]
    \centering
    \includegraphics[width=1\linewidth]{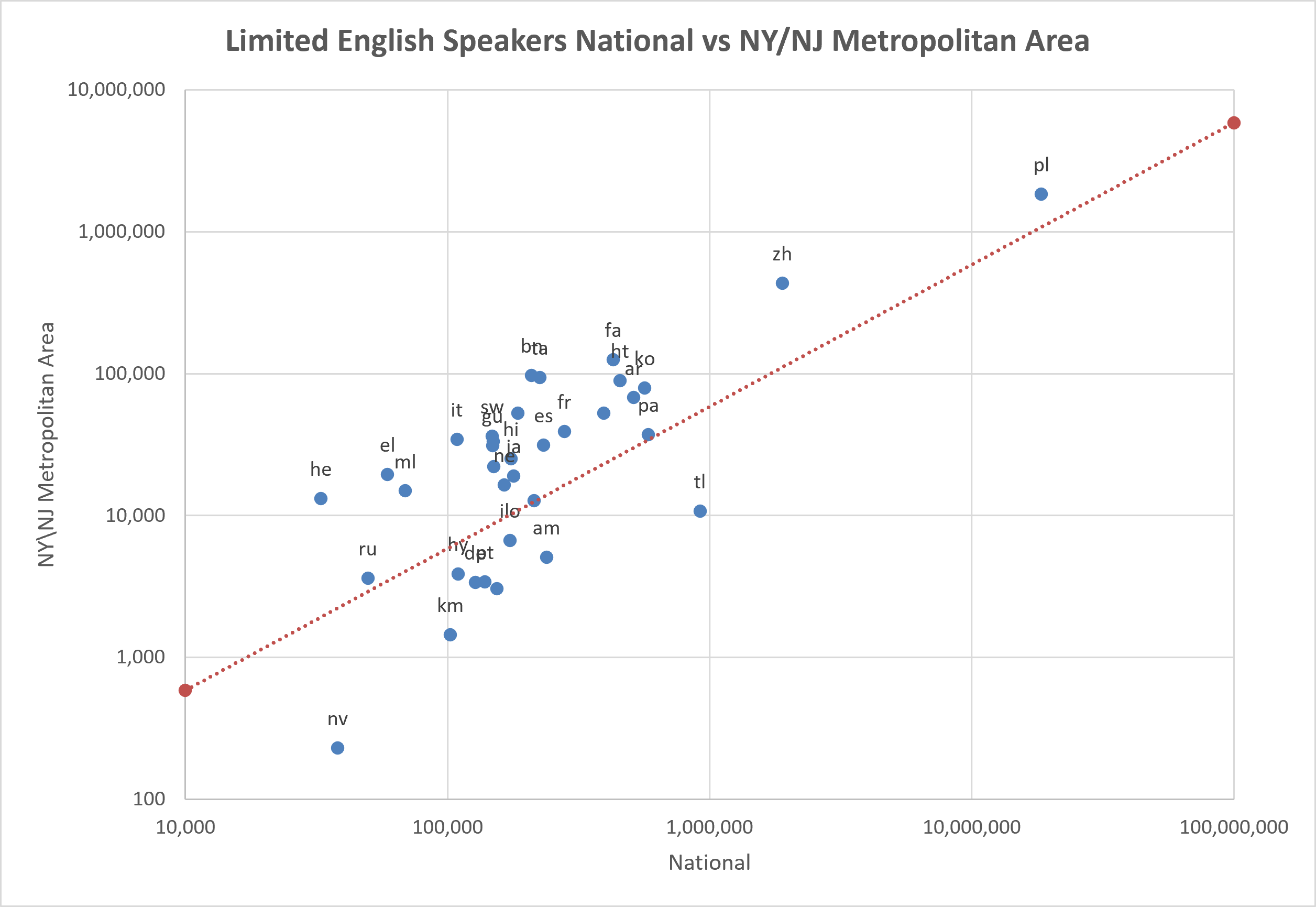}
    \caption{Count of Language Speakers in the NY/NJ Metropolitan area vs Nationally. Trend line represents the expected population if the percentages were identical in the NY/NJ Metropolitan area vs Nationally. \cite{Census2024ACSDT1Y2024.B16001}}
    \label{fig:translation_needs}
\end{figure}

The language needs of each geographic area vary because populations are not evenly distributed across the nation. Figure \ref{fig:translation_needs} shows the difference between the national and NY/NJ populations who speak a language other than English as their primary language. The population of NY/NJ is approximately 5.85\% of the total population of the United States. The trend line is the expected result if the percentages in NY/NJ were the same nationally. Languages that appear above the line are more heavily represented in NY/NJ while languages that appear below the line are more heavily represented nationally.

\label{appendix:translation_needs}

\input{tables/metropolitan_languages}

This data allows us to see some striking outliers in the populations. There are approximately 435,000 people who speak Chinese (zh) and ``Speak English Less than Well'' in NY/NJ accounting for nearly 23\% of the approximately 1,895,000 people nationally, much higher than the expected 5.85\%.  

Table \ref{tab:metropolitan_language_variation} reports a similar analysis across major metropolitan areas in the U.S, where the top row lists expected demographic share based on national population proportions. For example, Seattle-Tacoma-Bellevue, WA Metropolitan Area represents 1.22\% of the U.S. population, but contains substantial demographic skew: eighteen language communities are over-represented (>2 x baseline) and eight are under-represented (<0.5 x baseline).

We can use Haitian Creole\footnote{The US Census uses the language name Haitian.}
%Krey\`ol 
as a specific language example. In Table \ref{tab:metropolitan_language_variation} we see that it is under-represented in nearly every area. This is because Haitian Creole
%Krey\`ol 
speakers are disproportionally represented in Florida, were approximately 50\% of Haitian Creole
%Krey\`ol 
speakers live.

The data also shows that there is no language group within these areas that is spoken by the expected proportion of speakers. Each county, metropolitan area, or state has its own unique mix of languages, thus each area must to do their own language analysis to determine their community's language and communication needs.

\section{Languages and Language Varieties}
\label{app:languages}
If they chose, survey participants were able to provide the languages spoken by the communities they serve. The list below is an example of the data they shared. If participants shared a language variety (as they defined it), it is included here. The data is formatted showing the principal language, followed by varieties (if provided), \eg  \textbf{Language}: variety.\\
\\
\input{tables/laaww_languages}

\section{More on Post-editing and Evaluating Translations}
\label{appendix:more_on_evaluations}

Here are some rules of thumb for determining the adequacy of the translations provided by a commercial translation tool (or of any translations):
\begin{itemize}[noitemsep, topsep=0pt]
\item{} Use Multidimensional Quality Metric (MQM) \citep{mqm} to define what  \textit{major errors} might exist. Ultimately, there should be no \textit{major errors}, but minor errors may be acceptable.\footnote{Examples of major errors may include incorrectly translated safety instructions, incorrect locations, or incorrect times.}
\item{} Critical data, such as place names, should be kept in the original language because geographic place names tend to stay consistent and do not change based on language spoken. This information can be evaluated even by those who do not read the target language.
\end{itemize}

Post-editing is often done using Computer Assisted Translation (CAT) tools such as Trados or MemoQ\footnote{https://www.trados.com/, https://www.memoq.com/}, although it can be done in tools such as Word or Excel (with a little more difficulty). One of the advantages of using CAT tools is that they track translations and can reuse material from earlier translations (through \textit{Translation Memories}).  Many professional translators are familiar with these tools.

In scenarios where time is limited or it is not possible to do a formal evaluation, one can do round-trip translation and cross checking. For round-trip translation, use an MT tool to perform the initial translation into the target language. Once the translation output is generated, copy the translated text into a different MT tool and translate it back into the original language. If the new translation preserves the meaning, then the translation is likely usable. To use cross checking, use two different MT tools to produce a translation for the same text and compare the outputs. If the outputs are identical and likewise pass the back translation test, then the translation is likely usable.

There are many other aspects of translation evaluation, including fluency, tone, meaning, terminology, additions/omissions, cultural appropriateness, localization, accessibility, etc. These are beyond the scope of this paper.

\section{Survey Questions}
\label{app:survey_questions}
\input{Sections/Survey_Questions.tex}

\section{Evaluation Results}
\label{app:evaluation_results}
\input{tables/overall_v_na_languages}

\input{tables/chrf2026_overall}

%% file: tables/respondants.tex
\begin{table}[H]
\resizebox{\columnwidth}{!}{
\begin{tabular}{lcccc}\toprule
Type of EM & ID & Juridisction Scope & Location \\\midrule

Local Emergency Manager & \colorbox{LightGreen}{L1} & County & West Coast \\
Public Health Coordinator & \colorbox{LightPink}{H1} & County & West Coast \\
Public Information Officer & \colorbox{LightCyan}{P1} & City & East Coast \\
Advocate (Retired Researcher) & \colorbox{MediumTurquoise}{R1} & State & West Coast \\

\bottomrule
\end{tabular}
}
\caption{Demographic information of quoted survey participants.}
\label{tab:participant_information}
\end{table}

%% file: tables/metropolitan_languages.tex
\begin{table*}[!tb]
\resizebox{0.75\textwidth}{!}{\begin{minipage}{\textwidth}
\begin{tabular}{|l|r|rrrrrrr|} 
\hline

\multicolumn{1}{|l|}{} & \multicolumn{1}{c|}{National} & \multicolumn{1}{l|}{Atlanta} & \multicolumn{1}{l|}{Chicago} & \multicolumn{1}{l|}{Dallas} & \multicolumn{1}{l|}{Los Angeles} & \multicolumn{1}{l|}{Minneapolis}& \multicolumn{1}{l|}{San Francisco} & \multicolumn{1}{l|}{Seattle} \\ \hline
\multicolumn{1}{|l|}{Total} & \multicolumn{1}{c|}{321,745,943} & \multicolumn{1}{r|}{1.88\%} & \multicolumn{1}{r|}{2.77\%} & \multicolumn{1}{r|}{2.43\%} & \multicolumn{1}{r|}{3.82\%} & \multicolumn{1}{r|}{1.10\%} & \multicolumn{1}{r|}{1.37\%} & \multicolumn{1}{r|}{1.22\%} \\ \hline\hline
\begin{tabular}[c]{@{}l@{}}Amharic, Somali, or \\ other Afro-Asiatic languages\end{tabular} & 238,986 & 3.73\% & 3.56\% & \cellcolor[HTML]{92CDDC}4.96\% & 2.26\% & \cellcolor[HTML]{92CDDC}14.53\% & 1.84\% & \cellcolor[HTML]{92CDDC}6.31\% \\
\hline
Arabic & 513,293 & \cellcolor[HTML]{DDD9C4}0.48\% & 4.20\% & 3.23\% & 5.14\% & 0.65\% & 1.86\% & 1.11\% 
\\ \hline
Armenian & 109,630 & \cellcolor[HTML]{DDD9C4}0.00\% & \cellcolor[HTML]{DDD9C4}0.54\% & \cellcolor[HTML]{DDD9C4}0.00\% & \cellcolor[HTML]{92CDDC}83.24\% & \cellcolor[HTML]{DDD9C4}0.00\% & \cellcolor[HTML]{DDD9C4}0.40\% & 0.65\% \\
\hline
Bengali & 209,543 & 1.66\% & 1.74\% & 2.17\% & 3.35\% & \cellcolor[HTML]{DDD9C4}0.32\% & \cellcolor[HTML]{DDD9C4}0.17\% & 0.71\% \\
\hline
Chinese  & 1,895,001 & 1.36\% & 2.76\% & 1.33\% & \cellcolor[HTML]{92CDDC}14.23\% & \cellcolor[HTML]{DDD9C4}0.54\% & \cellcolor[HTML]{92CDDC}10.76\% & \cellcolor[HTML]{92CDDC}3.22\% \\
\hline
French   (incl. Cajun) & 279,230 & 2.76\% & 2.01\% & 3.11\% & 2.36\% & 1.22\% & 1.38\% & 1.61\% \\
\hline
German & 127,650 & \cellcolor[HTML]{DDD9C4}0.83\% & 1.98\% & \cellcolor[HTML]{DDD9C4}0.96\% & \cellcolor[HTML]{DDD9C4}1.86\% & 2.07\% & 0.79\% & 1.31\% \\
\hline
Greek & 59,054 & \cellcolor[HTML]{DDD9C4}0.73\% & \cellcolor[HTML]{92CDDC}8.86\% & \cellcolor[HTML]{DDD9C4}0.10\% & 2.49\% & \cellcolor[HTML]{DDD9C4}0.00\% & 0.73\% & \cellcolor[HTML]{DDD9C4}0.28\% \\
\hline
Gujarati & 148,495 & \cellcolor[HTML]{92CDDC}5.28\% & \cellcolor[HTML]{92CDDC}12.67\% & 3.60\% & 2.88\% & \cellcolor[HTML]{DDD9C4}0.48\% & 1.35\% & \cellcolor[HTML]{DDD9C4}0.45\% \\
\hline
Haitian & 454,970 & 1.73\% & \cellcolor[HTML]{DDD9C4}0.18\% & \cellcolor[HTML]{DDD9C4}0.21\% & \cellcolor[HTML]{DDD9C4}0.13\% & \cellcolor[HTML]{DDD9C4}0.26\% & \cellcolor[HTML]{DDD9C4}0.03\% & \cellcolor[HTML]{DDD9C4}0.10\% \\
\hline
Hebrew & 32,924 & 1.66\% & 1.64\% & \cellcolor[HTML]{DDD9C4}0.60\% & \cellcolor[HTML]{92CDDC}9.32\% & \cellcolor[HTML]{DDD9C4}0.29\% & 1.25\% & 2.37\% \\
\hline
Hindi & 174,630 & 3.09\% & 4.63\% & 4.39\% & 4.79\% & \cellcolor[HTML]{DDD9C4}0.45\% & \cellcolor[HTML]{92CDDC}7.12\% & \cellcolor[HTML]{92CDDC}2.66\% \\
\hline
Hmong & 91,284 & \cellcolor[HTML]{DDD9C4}0.69\% & \cellcolor[HTML]{DDD9C4}0.09\% & \cellcolor[HTML]{DDD9C4}0.38\% & \cellcolor[HTML]{DDD9C4}0.69\% & \cellcolor[HTML]{92CDDC}28.45\% & \cellcolor[HTML]{DDD9C4}0.00\% & \cellcolor[HTML]{DDD9C4}0.19\% \\
\hline
\begin{tabular}[c]{@{}l@{}}Ilocano, Samoan, Hawaiian, or \\ other Austronesian languages\end{tabular} & 173,239 & 1.06\% & 2.40\% & \cellcolor[HTML]{DDD9C4}1.09\% & 6.79\% & \cellcolor[HTML]{DDD9C4}0.25\% & \cellcolor[HTML]{92CDDC}3.78\% & \cellcolor[HTML]{92CDDC}4.88\% \\
\hline
Italian & 108,644 & \cellcolor[HTML]{DDD9C4}0.52\% & 5.52\% & 1.42\% & 2.43\% & \cellcolor[HTML]{DDD9C4}0.35\% & 1.55\% & \cellcolor[HTML]{DDD9C4}0.08\% \\
\hline
Japanese & 179,098 & 1.20\% & 2.39\% & 1.51\% & \cellcolor[HTML]{92CDDC}13.42\% & \cellcolor[HTML]{DDD9C4}0.54\% & \cellcolor[HTML]{92CDDC}4.20\% & \cellcolor[HTML]{92CDDC}3.72\% \\
\hline
Khmer & 102,600 & 2.19\% & \cellcolor[HTML]{DDD9C4}0.75\% & 2.36\% & \cellcolor[HTML]{92CDDC}14.97\% & \cellcolor[HTML]{92CDDC}2.36\% & \cellcolor[HTML]{92CDDC}3.74\% & \cellcolor[HTML]{92CDDC}7.49\% \\
\hline
Korean & 565,989 & \cellcolor[HTML]{92CDDC}4.23\% & 3.60\% & 3.16\% & \cellcolor[HTML]{92CDDC}22.66\% & \cellcolor[HTML]{DDD9C4}0.32\% & 2.30\% & \cellcolor[HTML]{92CDDC}4.41\% \\
\hline
\begin{tabular}[c]{@{}l@{}}Malayalam, Kannada, or \\ other Dravidian languages\end{tabular} & 69,100 & 1.23\% & \cellcolor[HTML]{92CDDC}6.26\% & \cellcolor[HTML]{92CDDC}12.09\% & \cellcolor[HTML]{DDD9C4}1.32\% & 0.73\% & 1.99\% & 2.43\% \\
\hline
Navajo & 38,127 & \cellcolor[HTML]{DDD9C4}0.00\% & \cellcolor[HTML]{DDD9C4}0.00\% & \cellcolor[HTML]{DDD9C4}0.30\% & \cellcolor[HTML]{DDD9C4}0.00\% & \cellcolor[HTML]{DDD9C4}0.03\% & \cellcolor[HTML]{DDD9C4}0.00\% & \cellcolor[HTML]{DDD9C4}0.12\% \\
\hline
\begin{tabular}[c]{@{}l@{}}Nepali, Marathi, or \\ other Indic languages\end{tabular} & 164,869 & 2.94\% & 1.51\% & \cellcolor[HTML]{92CDDC}7.76\% & 3.21\% & 0.64\% & \cellcolor[HTML]{92CDDC}4.47\% & 1.40\% \\
\hline
Persian   (incl. Farsi, Dari) & 213,566 & 2.43\% & 1.45\% & 2.27\% & \cellcolor[HTML]{92CDDC}20.99\% & \cellcolor[HTML]{DDD9C4}0.37\% & \cellcolor[HTML]{92CDDC}5.89\% & \cellcolor[HTML]{92CDDC}5.02\% \\
\hline
Polish & 185,755 & \cellcolor[HTML]{DDD9C4}0.37\% & \cellcolor[HTML]{92CDDC}35.58\% & \cellcolor[HTML]{DDD9C4}0.28\% & \cellcolor[HTML]{DDD9C4}0.99\% & \cellcolor[HTML]{DDD9C4}0.21\% & \cellcolor[HTML]{DDD9C4}0.37\% & \cellcolor[HTML]{DDD9C4}0.48\% \\
\hline
Portuguese & 394,010 & 1.63\% & \cellcolor[HTML]{DDD9C4}0.53\% & \cellcolor[HTML]{DDD9C4}1.20\% & \cellcolor[HTML]{DDD9C4}1.26\% & \cellcolor[HTML]{DDD9C4}0.18\% & 1.05\% & 1.22\% \\
\hline
Punjabi & 150,447 & \cellcolor[HTML]{DDD9C4}0.31\% & 2.01\% & \cellcolor[HTML]{DDD9C4}1.11\% & 2.46\% & \cellcolor[HTML]{DDD9C4}0.22\% & \cellcolor[HTML]{92CDDC}5.77\% & \cellcolor[HTML]{92CDDC}4.43\% \\
\hline
Russian & 429,378 & 1.01\% & 4.26\% & \cellcolor[HTML]{DDD9C4}0.57\% & 7.49\% & 1.32\% & 2.25\% & \cellcolor[HTML]{92CDDC}3.48\% \\
\hline
Serbo-Croatian & 79,111 & 1.93\% & \cellcolor[HTML]{92CDDC}13.69\% & 1.77\% & 2.22\% & 0.85\% & 0.72\% & 2.13\% \\
\hline
Spanish & 18,432,221 & 1.41\% & 3.46\% & 3.73\% & \cellcolor[HTML]{92CDDC}9.51\% & \cellcolor[HTML]{DDD9C4}0.39\% & 1.71\% & 0.62\% \\
\hline
\begin{tabular}[c]{@{}l@{}}Swahili or other languages \\ of Central, Eastern, and \\ Southern Africa\end{tabular} & 138,888 & \cellcolor[HTML]{92CDDC}7.02\% & \cellcolor[HTML]{DDD9C4}0.97\% & \cellcolor[HTML]{92CDDC}8.13\% & \cellcolor[HTML]{DDD9C4}0.84\% & \cellcolor[HTML]{92CDDC}2.94\% & \cellcolor[HTML]{DDD9C4}0.25\% & \cellcolor[HTML]{92CDDC}3.17\% \\
\hline
Tagalog   (incl. Filipino) & 584,724 & \cellcolor[HTML]{DDD9C4}0.49\% & 3.09\% & 1.40\% & \cellcolor[HTML]{92CDDC}14.52\% & 0.58\% & \cellcolor[HTML]{92CDDC}9.23\% & \cellcolor[HTML]{92CDDC}2.85\% \\
\hline
Tamil & 49,660 & 3.16\% & 3.91\% & \cellcolor[HTML]{92CDDC}5.99\% & 2.76\% & 1.25\% & \cellcolor[HTML]{92CDDC}5.10\% & \cellcolor[HTML]{92CDDC}3.43\% \\
\hline
Telugu & 104,199 & 3.69\% & 2.32\% & \cellcolor[HTML]{92CDDC}12.31\% & \cellcolor[HTML]{DDD9C4}0.84\% & 0.82\% & \cellcolor[HTML]{92CDDC}3.73\% & \cellcolor[HTML]{92CDDC}3.69\% \\
\hline
\begin{tabular}[c]{@{}l@{}}Thai, Lao, or \\other Tai-Kadai languages\end{tabular} & 154,016 & 1.17\% & 2.80\% & 3.08\% & \cellcolor[HTML]{92CDDC}9.62\% & 1.73\% & \cellcolor[HTML]{92CDDC}3.13\% & \cellcolor[HTML]{92CDDC}4.25\% \\
\hline
\begin{tabular}[c]{@{}l@{}}Ukrainian   or \\other Slavic languages\end{tabular} & 232,585 & \cellcolor[HTML]{DDD9C4}0.67\% & \cellcolor[HTML]{92CDDC}15.75\% & \cellcolor[HTML]{DDD9C4}0.42\% & 2.69\% & 1.28\% & 1.03\% & \cellcolor[HTML]{92CDDC}7.71\% \\
\hline
Urdu & 148,178 & 1.97\% & \cellcolor[HTML]{92CDDC}9.39\% & \cellcolor[HTML]{92CDDC}6.34\% & 3.96\% & \cellcolor[HTML]{DDD9C4}0.09\% & 1.92\% & 1.04\% \\
\hline
Vietnamese & 918,210 & 3.23\% & \cellcolor[HTML]{DDD9C4}1.37\% & \cellcolor[HTML]{92CDDC}4.87\% & \cellcolor[HTML]{92CDDC}18.03\% & 0.84\% & \cellcolor[HTML]{92CDDC}3.11\% & \cellcolor[HTML]{92CDDC}3.93\% \\
\hline
\begin{tabular}[c]{@{}l@{}}Yiddish, Pennsylvania \\Dutch or other \\West Germanic languages\end{tabular} & 225,109 & \cellcolor[HTML]{DDD9C4}0.05\% & \cellcolor[HTML]{DDD9C4}0.21\% & \cellcolor[HTML]{DDD9C4}0.12\% & \cellcolor[HTML]{DDD9C4}0.17\% & \cellcolor[HTML]{DDD9C4}0.10\% & \cellcolor[HTML]{DDD9C4}0.00\% & \cellcolor[HTML]{DDD9C4}0.09\% \\
\hline
\begin{tabular}[c]{@{}l@{}}Yoruba, Twi, Igbo, \\or other languages of \\Western Africa\end{tabular} & 149,225 & 2.09\% & 4.83\% & 2.93\% & 2.08\% & \cellcolor[HTML]{92CDDC}4.24\% & 0.78\% & \cellcolor[HTML]{92CDDC}2.53\%
\\\hline
\end{tabular}
\end{minipage} }
\caption{Language Variation of actual vs. expected representation for select Metropolitan areas in the United States \cite{Census2024ACSDT1Y2024.B16001}. \colorbox{tantable}{underrepresented = expected/2} \colorbox{bluetable}{overrepresented = expected*2}  \\
Languages spoken by fewer than 10,000 people are grouped with other languages by the U.S. Census Bureau. More information can be found here: \url{https://usa.ipums.org/usa/volii/00language.shtml}.}
\label{tab:metropolitan_language_variation}

\end{table*}

%% file: tables/laaww_languages.tex
\label{tab:laaww_languages}
\textbf{Ahtna};  
\textbf{Alutiiq / Sugpiaq}; \textbf{Amdo Tibetan}; 
\textbf{Amharic}; 
\textbf{Arabic}: Algerian, Chadian, Classical, Egyptian, Gulf,
North Levantine, South Levantine, Yemeni; 
\textbf{Armenian (Western)}; 
\textbf{Aromanian}; \textbf{Ashanti}; \textbf{Avestan}; \textbf{Azeri};

\textbf{Bahamian Creole}; \textbf{Bambara}; 
\textbf{Barbadian Creole}; \textbf{Bavarian}; 
\textbf{Belarusian}; \textbf{Bengali}; \textbf{Bislama}; 
\textbf{Brazilian}; \textbf{Bukhori}; 
\textbf{Burmese (Intha)};

\textbf{Calabrese}; \textbf{Cambodian / Khmer}; \textbf{Cebuano}; 
\textbf{Chinese}: Cantonese, Mandarin, Simplified, Fujianese,
Hainanese (Southern Min), Hokkien, Hunanese, Han, Shanghainese,
Taishanese, Taiwanese, Wenzhounese; 
\textbf{Coptic}; \textbf{Crimean Tatar}; \textbf{Croatian};

\textbf{Dari}; \textbf{Deg Xinag}; \textbf{Dena'ina}; 
\textbf{Dutch}; \textbf{Ecuadorian Kichwa}; \textbf{Efik}; 
\textbf{English}: African-American, Australian, Canadian, Ghanaian,
Hawaiian, Indian, Irish, Italian, Jewish, Liberian, Native American,
New Zealand, Newfoundland, Nigerian, Pakistani, Philippine; 
\textbf{Ewe}; \textbf{Eyak}; \textbf{Fante}; \textbf{Farsi}; 
\textbf{Finnish}; \textbf{French}; \textbf{Fulani};

\textbf{Galician}; \textbf{Garifuna}; \textbf{Geg Albanian}; 
\textbf{Georgian}; 
\textbf{German}: Austrian, Gottscheerisch, Swabian; 
\textbf{Greek}: Cretan, Cypriot, Koine; 
\textbf{Grenadian Creole}; \textbf{Gujarati}; 
\textbf{Guyanese Creole}; \textbf{Gwich'in}; \textbf{Haida}; 
\textbf{Haitian Creole}; \textbf{Hakha Chin}; 
\textbf{Hamgyŏng Korean}: Joseonjok; \textbf{Hebrew}: Biblical; 
\textbf{Hindi}; \textbf{Hmong}; \textbf{Holikachuk}; 
\textbf{Hungarian};

\textbf{Icelandic}; \textbf{Igbo}; \textbf{Ilocano}; \textbf{Indonesian};
\textbf{Inupiaq}; \textbf{Irpino}; \textbf{Italian}; 
\textbf{Iu Mien}; \textbf{Jamaican Patois}; \textbf{Japanese}; 
\textbf{Jewish Babylonian Aramaic}; \textbf{Judeo-Shirazi}; 
\textbf{Judeo-Spanish}; \textbf{Juhuri}; 
\textbf{Kalderash Romani}; \textbf{Kannada}; \textbf{Karen}; 
\textbf{Kinyarwanda}; \textbf{Kirundi}; \textbf{Konkanoi}; 
\textbf{Korean}; \textbf{Koyukon};

\textbf{Lahu}; \textbf{Lao}; \textbf{Lemko}; \textbf{Lenape}; 
\textbf{Lithuanian}; \textbf{Lukumí}; \textbf{Maithili}; 
\textbf{Malayalam}; \textbf{Maltese}; \textbf{Mandinka}; 
\textbf{Maranao}; \textbf{Marathi}; \textbf{Marshallese}; 
\textbf{Marwari}; \textbf{Matinecock}; \textbf{Mixtec}; 
\textbf{Mohican}; \textbf{Mon}; \textbf{Montaukett}; 
\textbf{Montserrat Creole};

\textbf{Nahuatl}; \textbf{Nauruan}; \textbf{Navajo}; 
\textbf{Ndebele}; \textbf{Neapolitan}; \textbf{Nepali}; 
\textbf{Newari}; \textbf{Noakhailla}; \textbf{North Frisian}; 
\textbf{Nuer}; \textbf{Pashto}; \textbf{Polish}; 
\textbf{Portuguese}: Brazilian, Portuguese; \textbf{Pugliese}; 
\textbf{Punjabi}; \textbf{Quiripi}; \textbf{Ramaluk}; 
\textbf{Romagnol}; \textbf{Romanian}; \textbf{Russian};

\textbf{Sandwippa}; \textbf{Sanskrit}; \textbf{Serer}; 
\textbf{Sherpa}; \textbf{Sicilian}; \textbf{Sinhalese}; 
\textbf{Somali}; \textbf{Soninke}; 
\textbf{Spanish}: Argentine, Bolivian, Castilian, Central American,
Colombian, Cuban, Dominican, Ecuadorian, Mexican, Puerto Rican,
Venezuelan; \textbf{Spanglish}; \textbf{St. Lawrence Island Yupik}; 
\textbf{St. Lucian Creole}; \textbf{Swahili}; \textbf{Sylheti};

\textbf{Tagalog}; \textbf{Tajik}; \textbf{Tamang}; \textbf{Tamil}; 
\textbf{Tanacross}; \textbf{Tanana}; \textbf{Telugu}; \textbf{Tetum}; 
\textbf{Thai}; \textbf{Tibetan}: Classical, Tö; \textbf{Tlingit}; 
\textbf{Tosk Albanian}; \textbf{Trinidadian Creole}; 
\textbf{Tsimshian}; \textbf{Turkish}; \textbf{Twi}; 
\textbf{Ukrainian}; \textbf{Unangam Tunuu / Aleut}; 
\textbf{Upper Kuskokwim}; \textbf{Upper Tanana}; \textbf{Urdu}; 
\textbf{Uzbek}; \textbf{Yiddish}: Hasidic, Litvish; 
\textbf{Yup'ik}; \textbf{Zulu}.

%% file: Sections/Survey_Questions.tex
The survey was presented to LAAWW membership during a monthly LAAWW meeting. Survey goals and methodologies were explained to the group and time was given for questions.

\noindent The survey questions below are reproduced as administered. Response choices and formats are provided. If \textit{Other} was a Response Choice, participants were able to use a Free Response to answer.

\begin{enumerate}
  \item \textbf{Which of the following best describes your current role(s) in emergency management?} \textbf{Response format:} Multiple Choice; Text \\
  \textbf{Response choices:} Local emergency manager; State emergency manager; Public information officer; Language coordinator / interpreter; Operations, logistics, or communications manager; Researcher; Other

  \item \textbf{Please list the city, county, or region your agency or institution primarily serves.} \\
  \textbf{Response format:} Free response

  \item \textbf{Approximately how many different languages are spoken by the communities you serve?} \\
  \textbf{Response format:}  Free response

  \item \textbf{If you have a convenient list of the languages, please paste it here or share a URL.} \\
  \textbf{Response format:}  Free response

  \item \textbf{Do you translate FROM languages other than English?} \\
  \textbf{Response format:}  Yes-no style response \\
  \textbf{Response choices:}  Yes; No

  \item \textbf{What languages do you translate FROM other than English?} \\
  \textbf{Response format:}  Free response

  \item \textbf{Feel free to share more information about any of your answers from this section.} \\
  \textbf{Response format:}  Free response

  \item \textbf{How does your agency/institution identify the languages to prioritize for translation and outreach?} \\
  \textbf{Response format:}  Multiple Choice; Text \\
  \textbf{Response choices:}  US Census or American Community Survey data; Your State's Department of Education Data; Local community or needs assessments; Input from schools, health, or social service agencies; Feedback from community organizations; Staff or volunteer knowledge; Other

  \item \textbf{What are the translation tools used by your agency/institution?} \\
  \textbf{Response format:}  Multiple Choice; Text \\
  \textbf{Response choices:}  Google Translate; Microsoft Translator; DeepL; ChatGPT; DeepSeek; Google Gemini; Staff member; Contractor; Community member or volunteer;  Other

  \item \textbf{What are the main reasons your agency/institution selected these tools or methods?} \\
  \textbf{Response format:}  Multiple Choice; Text \\
  \textbf{Response choices:}  Cost or budget constraints; Availability or ease of access; Accuracy or quality of translation; Easy to use or staff comfort; Compliance or policy requirements; Other

  \item \textbf{Has your agency/institution evaluated how well these tools work for accuracy, usability, or effectiveness?} \\
  \textbf{Response format:}  Yes-no style response \\
  \textbf{Response choices:}  Yes - formal evaluation; Yes - informal evaluation; No

  \item \textbf{Does your agency/institution review or update its translation tools or methods?} \\
  \textbf{Response format:}  Yes-no style response \\
  \textbf{Response choices:}  Yes - annually; Yes - as needed; No formal review process; Not sure; No

  \item \textbf{Who usually handles translations or multilingual communication within your agency/institution?} \\
  \textbf{Response format:}  Multiple Choice; Text \\
  \textbf{Response choices:}  Dedicated translation team; Individual staff member; Shared among staff; External vendor or contractor; Volunteers / community partners; Other

  \item \textbf{Feel free to share more information about tools, processes, personnel, and capacity here.} \\
  \textbf{Response format:}  Free response

  \item \textbf{What additional tools, training, or partnerships would most help your agency improve multilingual communication and translation capacity?} \\
  \textbf{Response format:}  Free response

  \item \textbf{What are some challenges?} \\
  \textbf{Response format:}  Free response

  \item \textbf{May we contact you if we have follow up questions? If so, please provide your preferred contact information.} \\
  \textbf{Response format:}  Free response
\end{enumerate}

%% file: tables/overall_v_na_languages.tex
\begin{table*}[t]
\centering
\small
\setlength{\tabcolsep}{5pt}
\renewcommand{\arraystretch}{1.08}
\begin{tabular}{l *{5}{r}}
\toprule

\textbf{Language} & \textbf{Microsoft} & \textbf{Google} & \textbf{\shortstack{GPT-5.4-\\mini}} & \textbf{Gemini 2026} & \textbf{Claude} \\
\midrule
\rowcolor{secbg} \multicolumn{6}{l}{\textbf{Pivot}} \\
\midrule
Arabic          & \cellcolor{pos2}7.34  & \cellcolor{pos2}7.42  & \cellcolor{pos2}7.66  & \cellcolor{pos2}8.34  & \cellcolor{pos2}7.67  \\
Chinese         & \cellcolor{pos2}11.97  & \cellcolor{pos2}10.66 & \cellcolor{pos2}13.02 & \cellcolor{pos2}11.10  & \cellcolor{pos2}12.67  \\
French          & \cellcolor{pos2}5.67  & \cellcolor{pos2}4.85  & \cellcolor{pos2}6.03  & \cellcolor{pos2}4.89  & \cellcolor{pos2}5.47  \\
Hindi           & \cellcolor{pos2}13.16 & \cellcolor{pos2}9.28  & \cellcolor{pos2}10.33 & \cellcolor{pos2}8.34  & \cellcolor{pos2}12.24 \\
Indonesian      & \cellcolor{pos2}1.98  & \cellcolor{pos2}3.33  & \cellcolor{pos2}2.97  & \cellcolor{pos2}3.03  & \cellcolor{pos2}2.44  \\
Portuguese (BR) & \cellcolor{pos2}4.41  & \cellcolor{pos2}4.41  & \cellcolor{pos2}4.42  & \cellcolor{pos2}4.69  & \cellcolor{pos2}4.93  \\

Russian         & \cellcolor{pos2}6.66  & \cellcolor{pos2}8.06  & \cellcolor{pos2}6.68  & \cellcolor{pos2}6.97  & \cellcolor{pos2}6.70  \\
Spanish (LA)    & \cellcolor{pos2}4.04  & \cellcolor{pos2}6.14  & \cellcolor{pos2}3.94  & \cellcolor{pos2}4.03  & \cellcolor{pos2}3.75  \\
Swahili         & \cellcolor{neg2}-2.96 & \cellcolor{neg2}-3.13 & \cellcolor{neg2}-3.04 & \cellcolor{neg2}-1.54 & \cellcolor{neg2}-2.30 \\
\midrule
\rowcolor{secbg} \multicolumn{6}{l}{\textbf{Important}} \\
\midrule
Bengali         & \cellcolor{pos2}1.65  & \cellcolor{pos2}1.05  & \cellcolor{pos2}0.04  & \cellcolor{neg2}-0.66 & \cellcolor{neg2}-0.13 \\
Farsi           & \cellcolor{pos2}1.99  & \cellcolor{pos2}3.53  & \cellcolor{pos2}4.78  & \cellcolor{pos2}4.17  & \cellcolor{pos2}3.96  \\
Marathi         & \cellcolor{pos2}2.07  & \cellcolor{pos2}1.72  & \cellcolor{pos2}1.20  & \cellcolor{pos2}0.90  & \cellcolor{pos2}1.70  \\
Myanmar         & \cellcolor{pos2}1.95  & \cellcolor{pos2}3.69  & \cellcolor{pos2}2.22  & \cellcolor{pos2}2.72  & \cellcolor{pos2}3.88  \\
Tagalog         & \cellcolor{pos2}1.47  & \cellcolor{pos2}2.52  & \cellcolor{pos2}2.13  & \cellcolor{pos2}2.40  & \cellcolor{pos2}0.75  \\
Tamil           & \cellcolor{pos2}0.03  & \cellcolor{pos2}0.28  & \cellcolor{neg2}-0.78 & \cellcolor{pos2}0.98  & \cellcolor{pos2}0.25  \\
Urdu            & \cellcolor{pos2}1.34  & \cellcolor{pos2}2.96  & \cellcolor{pos2}2.56  & \cellcolor{pos2}2.59  & \cellcolor{pos2}2.04  \\
\midrule
\rowcolor{secbg} \multicolumn{6}{l}{\textbf{Priority}} \\
\midrule
Amharic         & \cellcolor{pos2}4.61  & \cellcolor{pos2}4.47  & \cellcolor{pos2}1.96  & \cellcolor{pos2}3.73  & \cellcolor{pos2}1.41  \\
Dari            & \cellcolor{pos2}1.25  & --                    & \cellcolor{pos2}1.18  & \cellcolor{pos2}0.80  & \cellcolor{pos2}2.29  \\
Khmer           & \cellcolor{neg2}-6.54 & \cellcolor{neg2}-3.96 & \cellcolor{neg2}-3.40 & \cellcolor{neg2}-1.36 & \cellcolor{neg2}-2.54 \\
Kinyarwanda     & \cellcolor{pos2}1.32  & \cellcolor{pos2}0.03  & \cellcolor{neg2}-0.39 & \cellcolor{neg2}-1.08 & \cellcolor{pos2}0.60  \\
Nepali          & \cellcolor{pos2}3.45  & \cellcolor{pos2}3.91  & \cellcolor{pos2}4.65  & \cellcolor{pos2}4.35  & \cellcolor{pos2}3.76  \\
Pashto          & \cellcolor{pos2}5.94  & \cellcolor{pos2}10.11 & \cellcolor{pos2}5.34  & \cellcolor{pos2}2.98  & \cellcolor{pos2}4.96  \\
Somali          & \cellcolor{neg2}-0.21 & \cellcolor{neg2}-0.23 & \cellcolor{neg2}-1.01 & \cellcolor{pos2}0.07  & \cellcolor{neg2}-0.25 \\
Zulu            & \cellcolor{neg2}-2.19 & \cellcolor{neg2}-2.93 & \cellcolor{neg2}-1.38 & \cellcolor{neg2}-2.17 & \cellcolor{neg2}-0.59 \\
\bottomrule
\end{tabular}

\caption{Differences between Overall CHRF++ Score and News+Announcements CHRF++. \\\colorbox{pos2}{Blue} indicates Overall score is higher than N+A, \colorbox{neg2}{orange} indicates Overall is lower than N+A. Google Translate results were unavailable for Dari. Khmer, Myanmar, and Chinese do not use whitespace to mark word boundaries and CHRF was used instead. \cite{bansal-etal-2026}}
\label{tab:chrf_diff_scores_colored}
\end{table*}

%% file: tables/chrf2026_overall.tex
\begin{table*}[t]
\centering
\small
\setlength{\tabcolsep}{3.5pt}
\renewcommand{\arraystretch}{1.08}
\begin{tabular}{@{}l
  rr@{\hspace{8pt}}
  rr@{\hspace{8pt}}
  rr@{\hspace{8pt}}
  rr@{\hspace{8pt}}
  rr@{}}
\toprule
& \multicolumn{2}{c}{\textbf{Microsoft}}
& \multicolumn{2}{c}{\textbf{Google}}
& \multicolumn{2}{c}{\textbf{GPT-5.4-mini}}
& \multicolumn{2}{c}{\textbf{Gemini 2026}}
& \multicolumn{2}{c}{\textbf{Claude}} \\

\cmidrule(lr){2-3}
\cmidrule(lr){4-5}
\cmidrule(lr){6-7}
\cmidrule(lr){8-9}
\cmidrule(lr){10-11}
\textbf{Language}
& \textbf{Overall} & \textbf{N+A}
& \textbf{Overall} & \textbf{N+A}
& \textbf{Overall} & \textbf{N+A}
& \textbf{Overall} & \textbf{N+A}
& \textbf{Overall} & \textbf{N+A} \\
\midrule
\rowcolor{secbg}
\multicolumn{11}{@{}l@{}}{\textbf{Pivot}} \\
\midrule
Arabic
& \cellcolor{scoremid}55.08 & \cellcolor{scoremid}47.73
& \cellcolor{scoremid}\textbf{59.23} & \cellcolor{scoremid}\textbf{51.82}
& \cellcolor{scoremid}57.86 & \cellcolor{scoremid}50.20
& \cellcolor{scoremid}57.27 & \cellcolor{scoremid}48.93
& \cellcolor{scoremid}55.64 & \cellcolor{scoremid}47.97 \\
Chinese
& \cellcolor{scoremid}45.74 & \cellcolor{scorelow}33.77
& \cellcolor{scoremid}\textbf{50.48} & \cellcolor{scorelow}\textbf{39.82}
& \cellcolor{scoremid}48.10 & \cellcolor{scorelow}35.08
& \cellcolor{scorelow}42.66 & \cellcolor{scorelow}31.56
& \cellcolor{scoremid}45.89 & \cellcolor{scorelow}33.21 \\
French
& \cellcolor{scorehigh}63.64 & \cellcolor{scoremid}57.97
& \cellcolor{scorehigh}63.13 & \cellcolor{scoremid}58.28
& \cellcolor{scorehigh}\textbf{65.52} & \cellcolor{scoremid}\textbf{59.49}
& \cellcolor{scorehigh}63.05 & \cellcolor{scoremid}58.16
& \cellcolor{scorehigh}63.02 & \cellcolor{scoremid}57.55 \\
Hindi
& \cellcolor{scorehigh}\textbf{65.62} & \cellcolor{scoremid}52.46
& \cellcolor{scorehigh}61.76 & \cellcolor{scoremid}52.48
& \cellcolor{scorehigh}62.81 & \cellcolor{scoremid}52.48
& \cellcolor{scorehigh}61.75 & \cellcolor{scoremid}\textbf{53.41}
& \cellcolor{scorehigh}62.42 & \cellcolor{scoremid}50.18 \\
Indonesian
& \cellcolor{scorehigh}74.72 & \cellcolor{scorehigh}\textbf{72.74}
& \cellcolor{scorehigh}74.88 & \cellcolor{scorehigh}71.54
& \cellcolor{scorehigh}\textbf{74.93} & \cellcolor{scorehigh}71.95
& \cellcolor{scorehigh}73.92 & \cellcolor{scorehigh}70.88
& \cellcolor{scorehigh}73.74 & \cellcolor{scorehigh}71.30 \\
Portuguese (BR)
& \cellcolor{scorehigh}72.06 & \cellcolor{scorehigh}67.65
& \cellcolor{scorehigh}72.05 & \cellcolor{scorehigh}67.64
& \cellcolor{scorehigh}\textbf{72.99} & \cellcolor{scorehigh}\textbf{68.56}
& \cellcolor{scorehigh}\textbf{72.99} & \cellcolor{scorehigh}68.30
& \cellcolor{scorehigh}72.62 & \cellcolor{scorehigh}67.69 \\
Russian
& \cellcolor{scoremid}56.13 & \cellcolor{scoremid}49.47
& \cellcolor{scoremid}\textbf{59.57} & \cellcolor{scoremid}\textbf{51.51}
& \cellcolor{scoremid}58.11 & \cellcolor{scoremid}51.43
& \cellcolor{scoremid}58.19 & \cellcolor{scoremid}51.23
& \cellcolor{scoremid}57.14 & \cellcolor{scoremid}50.44 \\
Spanish (LA)
& \cellcolor{scorehigh}71.88 & \cellcolor{scorehigh}67.84
& \cellcolor{scorehigh}\textbf{74.17} & \cellcolor{scoremid}68.03
& \cellcolor{scorehigh}74.08 & \cellcolor{scorehigh}\textbf{70.14}
& \cellcolor{scorehigh}73.64 & \cellcolor{scorehigh}69.60
& \cellcolor{scorehigh}72.84 & \cellcolor{scorehigh}69.09 \\

Swahili
& \cellcolor{scoremid}59.96 & \cellcolor{scorehigh}62.92
& \cellcolor{scorehigh}\textbf{61.59} & \cellcolor{scorehigh}\textbf{64.72}
& \cellcolor{scorehigh}60.44 & \cellcolor{scorehigh}63.48
& \cellcolor{scorehigh}60.16 & \cellcolor{scorehigh}61.70
& \cellcolor{scoremid}55.86 & \cellcolor{scoremid}58.16 \\
\midrule
\rowcolor{secbg}
\multicolumn{11}{@{}l@{}}{\textbf{Important}} \\
\midrule
Bengali
& \cellcolor{scoremid}\textbf{51.40} & \cellcolor{scoremid}\textbf{49.75}
& \cellcolor{scoremid}46.69 & \cellcolor{scoremid}45.64
& \cellcolor{scoremid}47.25 & \cellcolor{scoremid}47.21
& \cellcolor{scorelow}44.38 & \cellcolor{scoremid}45.04
& \cellcolor{scoremid}47.55 & \cellcolor{scoremid}47.68 \\
Farsi
& \cellcolor{scoremid}52.32 & \cellcolor{scoremid}50.33
& \cellcolor{scoremid}\textbf{56.44} & \cellcolor{scoremid}\textbf{52.92}
& \cellcolor{scoremid}52.71 & \cellcolor{scoremid}47.93
& \cellcolor{scoremid}51.80 & \cellcolor{scoremid}47.63
& \cellcolor{scoremid}51.38 & \cellcolor{scoremid}47.42 \\
Marathi
& \cellcolor{scoremid}\textbf{46.80} & \cellcolor{scorelow}\textbf{44.73}
& \cellcolor{scoremid}46.05 & \cellcolor{scorelow}44.33
& \cellcolor{scorelow}44.97 & \cellcolor{scorelow}43.76
& \cellcolor{scorelow}44.33 & \cellcolor{scorelow}43.43
& \cellcolor{scorelow}41.57 & \cellcolor{scorelow}39.87 \\
Myanmar
& \cellcolor{scoremid}48.16 & \cellcolor{scoremid}46.21
& \cellcolor{scoremid}\textbf{54.56} & \cellcolor{scoremid}50.87
& \cellcolor{scoremid}53.99 & \cellcolor{scoremid}\textbf{51.77}
& \cellcolor{scoremid}52.62 & \cellcolor{scoremid}49.90
& \cellcolor{scorelow}42.84 & \cellcolor{scorelow}38.95 \\
Tagalog
& \cellcolor{scorehigh}64.72 & \cellcolor{scorehigh}63.25
& \cellcolor{scorehigh}\textbf{68.27} & \cellcolor{scorehigh}\textbf{65.75}
& \cellcolor{scorehigh}66.61 & \cellcolor{scorehigh}64.48
& \cellcolor{scorehigh}63.42 & \cellcolor{scorehigh}61.02
& \cellcolor{scorehigh}63.83 & \cellcolor{scorehigh}63.08 \\
Tamil
& \cellcolor{scoremid}\textbf{56.87} & \cellcolor{scoremid}\textbf{56.85}
& \cellcolor{scoremid}53.11 & \cellcolor{scoremid}52.84
& \cellcolor{scoremid}48.75 & \cellcolor{scoremid}49.53
& \cellcolor{scoremid}50.68 & \cellcolor{scoremid}49.70
& \cellcolor{scoremid}46.48 & \cellcolor{scoremid}46.23 \\
Urdu
& \cellcolor{scoremid}50.77 & \cellcolor{scoremid}49.42
& \cellcolor{scoremid}\textbf{52.80} & \cellcolor{scoremid}\textbf{49.84}
& \cellcolor{scoremid}50.67 & \cellcolor{scoremid}48.12
& \cellcolor{scoremid}49.17 & \cellcolor{scoremid}46.58
& \cellcolor{scoremid}49.37 & \cellcolor{scoremid}47.33 \\
\midrule
\rowcolor{secbg}
\multicolumn{11}{@{}l@{}}{\textbf{Priority}} \\
\midrule
Amharic
& \cellcolor{scorelow}\textbf{35.84} & \cellcolor{scorelow}31.23
& \cellcolor{scorelow}35.78 & \cellcolor{scorelow}\textbf{31.31}
& \cellcolor{scorelow}31.23 & \cellcolor{scorelow}29.27
& \cellcolor{scorelow}32.62 & \cellcolor{scorelow}28.89
& \cellcolor{scorelow}27.60 & \cellcolor{scorelow}26.18 \\
Dari
& \cellcolor{scoremid}\textbf{48.15} & \cellcolor{scoremid}\textbf{46.90}
& -- & --
& \cellcolor{scorelow}44.23 & \cellcolor{scorelow}43.05
& \cellcolor{scorelow}44.56 & \cellcolor{scorelow}43.76
& \cellcolor{scorelow}43.63 & \cellcolor{scorelow}41.34 \\
Khmer
& \cellcolor{scorehigh}60.53 & \cellcolor{scorehigh}\textbf{67.07}
& \cellcolor{scorehigh}\textbf{61.60} & \cellcolor{scorehigh}65.66
& \cellcolor{scoremid}57.52 & \cellcolor{scorehigh}60.92
& \cellcolor{scoremid}56.86 & \cellcolor{scoremid}58.22
& \cellcolor{scoremid}46.66 & \cellcolor{scoremid}49.20 \\
Kinyarwanda
& \cellcolor{scoremid}45.88 & \cellcolor{scorelow}44.56
& \cellcolor{scoremid}\textbf{47.87} & \cellcolor{scoremid}47.84
& \cellcolor{scoremid}47.74 & \cellcolor{scoremid}48.13
& \cellcolor{scoremid}47.50 & \cellcolor{scoremid}\textbf{48.59}
& \cellcolor{scorelow}33.23 & \cellcolor{scorelow}32.64 \\
Nepali
& \cellcolor{scoremid}53.42 & \cellcolor{scoremid}49.98
& \cellcolor{scoremid}\textbf{56.19} & \cellcolor{scoremid}\textbf{52.27}
& \cellcolor{scoremid}52.39 & \cellcolor{scoremid}47.75
& \cellcolor{scoremid}49.35 & \cellcolor{scoremid}45.00
& \cellcolor{scoremid}50.25 & \cellcolor{scoremid}46.49 \\
Pashto
& \cellcolor{scorelow}39.47 & \cellcolor{scorelow}33.53
& \cellcolor{scoremid}\textbf{48.52} & \cellcolor{scorelow}\textbf{38.40}
& \cellcolor{scorelow}42.25 & \cellcolor{scorelow}36.91
& \cellcolor{scorelow}40.10 & \cellcolor{scorelow}37.12
& \cellcolor{scorelow}33.66 & \cellcolor{scorelow}28.70 \\
Somali
& \cellcolor{scorelow}44.92 & \cellcolor{scoremid}45.12
& \cellcolor{scoremid}\textbf{46.62} & \cellcolor{scoremid}\textbf{46.85}
& \cellcolor{scoremid}45.44 & \cellcolor{scoremid}46.46
& \cellcolor{scorelow}44.53 & \cellcolor{scorelow}44.46
& \cellcolor{scorelow}37.18 & \cellcolor{scorelow}37.44 \\
Zulu
& \cellcolor{scoremid}52.76 & \cellcolor{scoremid}54.95
& \cellcolor{scoremid}\textbf{53.61} & \cellcolor{scoremid}\textbf{56.54}
& \cellcolor{scoremid}53.34 & \cellcolor{scoremid}54.73
& \cellcolor{scoremid}50.31 & \cellcolor{scoremid}52.49
& \cellcolor{scorelow}42.54 & \cellcolor{scorelow}43.13 \\
\bottomrule
\end{tabular}
\caption{Overall and News+Announcements (N+A) CHRF++ scores for English$\rightarrow$Target Language across translation systems and languages. Highest scoring language result for Overall and N+A is in bold. Shading indicates relative score ranges within the evaluation: \colorbox{scorehigh}{blue} represents higher scores (>60), \colorbox{scoremid}{pale green} represents mid-range scores (45-60), and \colorbox{scorelow}{orange} represents lower scores (<45). CHRF++ does not have a universal safety or usability threshold, so these colors indicate relative performance only.  Google Translate results were unavailable for Dari. Khmer, Myanmar, and Chinese do not use whitespace to mark word boundaries and CHRF was used instead. \cite{bansal-etal-2026}}
\label{tab:chrf_scores_spaced}
\end{table*}